\documentclass[conference]{IEEEtran}
\IEEEoverridecommandlockouts

\usepackage{cite}
\usepackage{amsmath,amssymb,amsfonts}
\usepackage{algorithmic}
\usepackage{multirow}
\usepackage{pifont}
\usepackage[backref]{hyperref}
\usepackage{graphicx}
\usepackage{textcomp}
\usepackage{xcolor}
\hypersetup{
hidelinks}

\def\BibTeX{{\rm B\kern-.05em{\sc i\kern-.025em b}\kern-.08em
    T\kern-.1667em\lower.7ex\hbox{E}\kern-.125emX}}
\begin{document}

\title{Prototype-Driven Multi-Feature Generation for Visible-Infrared Person Re-identification\\}
\author{
\IEEEauthorblockN{Jiarui Li$^{1}$, Zhen Qiu$^{1}$, Yilin Yang$^{1}$, Yuqi Li$^{1}$\thanks{Jiarui Li, Zhen Qiu, Yilin Yang, Yuqi Li are interns}, Zeyu Dong$^{2}$, Chuanguang Yang$^{1}$\dag\thanks{\IEEEauthorrefmark{2}Corresponding author, Email: yangchuanguang@ict.ac.cn}}
\IEEEauthorblockA{\textit{$^1$Institute of Computing Technology, Chinese Academy of Sciences, China} \\
\textit{$^2$The art \& science college, Boston University, USA}
}
}

\setlength{\textfloatsep}{7pt}
\setlength{\intextsep}{7pt}
\maketitle
\begin{abstract}
The primary challenges in visible-infrared person re-identification arise from the differences between visible (\textit{vis}) and infrared (\textit{ir}) images, including inter-modal and intra-modal variations. These challenges are further complicated by varying viewpoints and irregular movements. Existing methods often rely on horizontal partitioning to align part-level features, which can introduce inaccuracies and have limited effectiveness in reducing modality discrepancies. In this paper, we propose a novel Prototype-Driven Multi-feature generation framework (PDM) aimed at mitigating cross-modal discrepancies by constructing diversified features and mining latent semantically similar features for modal alignment. PDM comprises two key components: Multi-Feature Generation Module (MFGM) and Prototype Learning Module (PLM). The MFGM generates diversity features closely distributed from modality-shared features to represent pedestrians. Additionally, the PLM utilizes learnable prototypes to excavate latent semantic similarities among local features between visible and infrared modalities, thereby facilitating cross-modal instance-level alignment. We introduce the cosine heterogeneity loss to enhance prototype diversity for extracting rich local features. Extensive experiments conducted on the SYSU-MM01 and LLCM datasets demonstrate that our approach achieves state-of-the-art performance. Our codes are available at \url{https://github.com/mmunhappy/ICASSP2025-PDM}.
\end{abstract}


\section{Introduction}
Person re-identification (ReID), a process of recognizing individuals across various image datasets taken by different cameras, commonly focuses on RGB images captured in ideal daylight conditions. This preference often leads to diminished effectiveness and unreliable outcomes in low-light or night-time environments. As a solution to this limitation, especially for continuous surveillance needs, the domain of visible-infrared person re-identification (VI-ReID) has emerged as a key area of research. The growing deployment of intelligent surveillance cameras, which can switch automatically to infrared mode, has further accelerated progress in this field.

VI-ReID~\cite{huang2023deep} presents a more complex challenge than traditional ReID. It must navigate not only intra-modality variances but also cross-modality differences that stem from the distinct imaging techniques of visible (VIS) and infrared (IR) cameras. Existing approaches~\cite{liu2020parameter, qi2023fine, fu2021cm} primarily focus on mapping VIS and IR features into a unified embedding space with the aim of minimizing cross-modality dissimilarities. Additionally, they attempt to address intra-modality variations – caused by changes in viewpoint, obstruction, and background – by segmenting body features horizontally and aligning them based on minimal feature distances. Nevertheless, such methods often neglect the dynamic positioning of body parts, leading to semantic misalignments that can impair the effectiveness of ReID.

Some approaches ~\cite{wang2020cross,wang2019rgb, zhang2018alignedreid} involve the use of Generative Adversarial Networks (GANs) to convert infrared or visible images into the opposite modality, thereby bridging the modality gap. However, these techniques are hampered by limited training data and the intrinsic noise in the image transformation process, affecting their overall efficacy.

In this paper, we propose a Prototype-Driven Multi-Feature Generation (PDM) framework designed to align modal features using two primary strategies: generating diverse features that closely match in distribution to minimize inter-modal disparities, and extracting semantically similar local features. The framework consists of a Multi-Feature Generation Module (MFGM) and a Prototype Learning Module (PLM). 

Specifically, the MFGM employs center-guided pair mining loss to generate diverse features, reducing modality differences and enriching the feature representation for PLM. The PLM assigns weights to modality features based on the similarity with learnable prototypes, thereby revealing latent semantically similar local features and achieving feature alignment. Furthermore, we introduce a dual-center separation loss to enhance the network's ability to discriminate pedestrian relationships.

Our contributions are twofold:

$\bullet$ We introduce a prototype-driven multi-feature generation framework, where the MFGM is utilized to generate diverse features that are distributed closely. The PLM module is responsible for mining local features by latent semantic similarity between VIS and IR modality features, thus achieving instance-level feature alignment.

$\bullet$  Extensive experiments conducted on the SYSU-MM01~\cite{wu2017rgb} and LLCM datasets demonstrate that the proposed method achieves state-of-the-art performance.

\section{Related Work}
Generally speaking, there are two main categories of methods in VI-ReID: the feature-level methods and the image-level methods.

Feature-level methods primarily focus on feature learning, aiming to minimize the disparity between distinct features and their common analogs in the feature space. For instance, MSCLNet \cite{zhang2022modality} bolsters the representation of modality-specific features through a cascaded amalgamation of modality cooperative complementary learning methods. Likewise, FIENet \cite{qi2023fine} engages intermediate features and undertakes fine-grained learning, anchored by identity-constrained feature centers. Despite their efficacy in enhancing performance, these methods tend to over-rely on global features, thereby neglecting vital local information, potentially leading to suboptimal results.

Conversely, techniques such as HCT \cite{liu2020parameter} and MAUM \cite{liu2022learning} address this issue by employing Part-based Convolutional Blocks (PCB) to directly extract features from horizontal partitions. This approach augments feature representation. Furthermore, HHRG \cite{feng2021homogeneous} develops a homograph between the component features of horizontal partitions and global features, promoting effective alignment of local features and further elevating saliency. However, the unpredictable movement of pedestrians may result in misalignment of horizontal component features, which could diminish the effectiveness of these methods.

Image-level methods primarily revolve around converting one modality into another to alleviate the cross-modality gap between Visible (VIS) and Infrared (IR) images. Techniques such as cmGAN and D2RL utilize Generative Adversarial Networks (GANs) to minimize these modality differences. AlignGAN \cite{wang2019rgb} employs GANs for aligning cross-modality features at both the pixel and feature levels, while FMCNet \cite{zhang2022fmcnet} implements feature-level modality compensation using GANs. Moreover, X-modality \cite{li2020infrared} and MMN \cite{zhang2021towards} introduce an intermediate modality to bridge the gap between VIS and IR feature distributions. Nonetheless, these methods still face challenges in effectively mitigating modality discrepancies.

\section{Method}
Motivated by the need to address key challenges in VI-ReID, we introduce PDM. Our approach aims to overcome limitations of existing methods that rely on constructing additional intermediate modality images. Instead, we focus on generating diverse yet closely distributed features to effectively represent pedestrians and bridge the modality gap. Inspired by prototype learning, we leverage learnable prototypes to extract semantically similar local features across modalities, facilitating modal instance-level alignment.

\begin{figure*}[t]
    \centering
    \includegraphics[width=2\columnwidth]
    {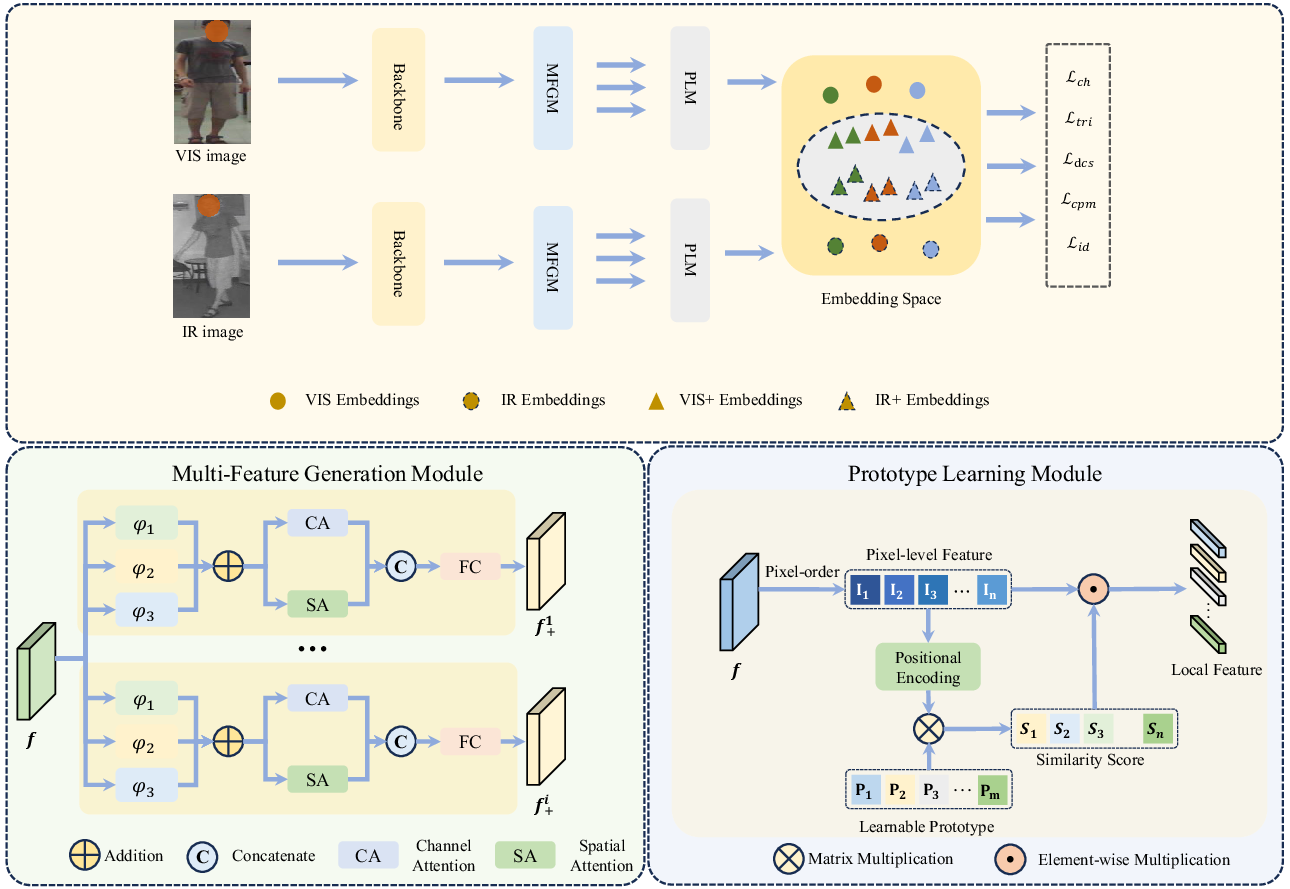}
    \caption{The Framework of PDM.}
    \label{fig:flowchart}
\end{figure*}

The network architecture of PDM is depicted in Fig.~\ref{fig:flowchart}, consisting of two primary components: the Multi-Feature Generation Module (MFGM) and the Prototype Learning Module (PLM). Initially, MFGM processes visual (VIS) and infrared (IR) features extracted by the backbone network to generate diverse yet closely distributed features. Subsequently, PLM extracts semantically similar local features across VIS and IR modalities. These combined local and global features are then utilized for pedestrian discrimination, guided by various loss functions during model training. 

\subsection{Multi-Feature Generation Module (MFGM)}\label{MFGM}
The MFGM consists of (i) identical branches, illustrated in Fig.~\ref{fig:flowchart}. Initially, the feature map (\(f\)) undergoes three \(3\times3\) dilated convolutions with dilation rates of 1, 2, and 3, respectively, to capture information from varying receptive fields. The outputs are then fused, reducing the channel dimension to one-fourth of its original size. To enhance non-linear representations, sequential operations include channel attention (\(\textbf{CA}\)), spatial attention (\(\textbf{SA}\)), and ReLU activation. A fully connected \((\mathcal{FC})\) layer aligns the channel dimension with the original feature map (\(f\)). The outputs \(f^{i}_{+}\) from all branches, along with \(f\), are concatenated to form the input for the next stage of the network. The resulting embeddings \(f^{i}_{+}\) for each branch are formulated as follows:

\begin{equation}
    {f}^{i} = (\mathbf{\phi}^1_{3\times3}(f) + \mathbf{\phi}^2_{3\times3}(f) + \mathbf{\phi}^3_{3\times3}(f))
\end{equation}

\begin{equation}
    f^{i}_{+} = \mathcal{FC}(\text{ReLU}([\textbf{CA}({f}^{i}),\textbf{SA}({f}^{i})]))
\end{equation}
where \([\cdot,\cdot]\) represents concatenation.

\textbf{Center-Guided Pair Mining Loss.} To enhance the diversity of the generated embeddings \(f^{i}_{+}\), we incorporate the center-guided pair mining loss \(\mathcal{L}_{cpm}\), following the DEEN~\cite{zhang2023diverse} approach. The \(\mathcal{L}_{cpm}\) for the \textbf{VIS} and \textbf{IR} modalities are defined as:

\begin{small}
\begin{equation}
\label{eq:vis_cpmloss}
\begin{aligned}
\mathcal{L}(\mathbf{c}_v, \mathbf{c}_{ir}, \mathbf{c}_{v+}^{i}) &= [\mathbf{\textit{D}}(\mathbf{c}^{j}_{ir}, \mathbf{c}^{i, j}_{v+}) - \mathbf{\textit{D}}(\mathbf{c}^{j}_{v}, \mathbf{c}^{i, j}_{v+}) \\
&\quad- \mathbf{\textit{D}}(\mathbf{c}^{j}_{v}, \mathbf{c}^{k}_{v}) + \alpha]_{+}.
\end{aligned}
\end{equation}
\end{small}

\begin{small}
\begin{equation}
\label{eq:ir_cpmloss}
\begin{aligned}
\mathcal{L}(\mathbf{c}_v, \mathbf{c}_{ir}, \mathbf{c}_{ir+}^{i}) &= [\mathbf{\textit{D}}(\mathbf{c}^{j}_{v}, \mathbf{c}^{i, j}_{ir+}) - \mathbf{\textit{D}}(\mathbf{c}^{j}_{ir}, \mathbf{c}^{i, j}_{ir+}) \\
&\quad- \mathbf{\textit{D}}(\mathbf{c}^{j}_{ir}, \mathbf{c}^{k}_{ir}) + \alpha]_{+}.
\end{aligned}
\end{equation}
\end{small}
where \(\textbf{\textit{D}}(\cdot, \cdot)\) denotes Euclidean distance. \(\mathbf{c}_v^i\) and \(\mathbf{c}_{ir}^i\) represent the original feature centers from VIS and IR modalities, while \(\mathbf{c}^i_{v+}\) and \(\mathbf{c}^i_{ir+}\) are the feature centers for generated embeddings \(f^{v+}\) and \(f^{ir+}\). Indices \(j\) and \(k\) denote distinct identities in a mini-batch, and \([\delta]_+ = \max(\delta, 0)\). The margin term \(\alpha\) is included for balanced optimization.

Therefore, the total \(\mathcal{L}_{cpm}\) can be formulated as:

\begin{equation}
    \mathcal{L}_{cpm} = \mathcal{L}(\mathbf{c}_v, \mathbf{c}_{ir}, \mathbf{c}^i_{v^+}) + \mathcal{L}(\mathbf{c}_v, \mathbf{c}_{ir}, \mathbf{c}^i_{{ir}^+})
\end{equation}

\subsection{Prototype Learning Module (PLM)}\label{PLM}
The PLM is illustrated in Fig.~\ref{fig:flowchart}, utilizing multiple learnable prototypes to extract semantically similar features from \( f^{\textit{v}} \) and \( f^{\textit{ir}} \), each represented in \( \mathbb{R}^{h \times w \times c} \), where \( h \), \( w \), and \( c \) denote the height, width, and channel dimensions of the feature maps. We adjust the weights of modality-specific features based on similarity scores between prototypes and features, where higher scores signify stronger semantic relevance. This adaptation enables PLM to effectively capture semantically similar local features. Specifically, we define a set of learnable prototypes \( \mathbf{P} = [\mathbf{P}_1, \mathbf{P}_2, \ldots, \mathbf{P}_m] \in \mathbb{R}^{m \times c} \) to encapsulate latent similar features, with \( \mathbf{P}_i \in \mathbb{R}^{1 \times c} \) representing the \( i \)-th prototype and \( m \) denoting the total number.


The process of extracting semantically similar local features using prototypes is consistent for both \( f^{\textit{v}} \) and \( f^{\textit{ir}} \). For the \( f^{\textit{v}} \), organized pixel-wise as \(\mathbf{I}_v = [\mathbf{I}^1_v, \mathbf{I}^2_v, \ldots, \mathbf{I}^n_v]\) in \(\mathbb{R}^{n \times c}\) with \(n = h \times w\), we incorporate position encoding for spatial consistency. The similarity between \(\mathbf{I}_v\) and \(\mathbf{P}\) is calculated, producing a similarity matrix \(\mathbf{S} \in \mathbb{R}^{m \times n}\), as described in Eq.~\ref{eq:Similary}.

\begin{equation}
    \mathbf{S} = \sigma\left(\mathbf{P} \otimes \mathbf{I}_v\right)
    \label{eq:Similary}
\end{equation}
where \(\otimes\) denotes matrix multiplication and \(\sigma(\cdot)\) represents the sigmoid activation function.

Subsequently, by weighting pixel-level features with \(\mathbf{S}\), we obtain semantically similar local features. The process can be described as follows:

\begin{equation}
    \mathbf{p}^i_v = \frac{1}{n} \sum_{i=1}^{n} (\mathbf{S}^{ij}_{v} \odot \mathbf{I}^{i}_{v})
    \label{eq:latent}
\end{equation}
where \(\odot\) represents element multiplication, and \(\mathbf{S}^{ij}_{v}\)  represents the similarity score between the \(i\)-th prototype and the \(j\)-th pixel.

Finally, we concatenate the \(\mathbf{p}^i_v\) with the global feature to obtain the final feature \(\mathbf{F}_{v} \in \mathbb{R}^{(m+1)c}\).

\begin{table*}[!t]
\centering
\caption{Comparison with cross-modality ReID methods on SYSU-MM01 and LLCM datasets. 1\textsuperscript{st} best results are in bold.}
\label{tab:dbsy}
 \fontsize{7}{9}\selectfont
\resizebox{1.95\columnwidth}{!}{%
\begin{tabular}{cc|cccc|cccc}
\hline
\multicolumn{2}{c|}{Datasets}                                                     & \multicolumn{4}{c|}{SYSU-MM01}                                                & \multicolumn{4}{c}{LLCM}                                                                                     \\ \hline
\multicolumn{2}{c|}{Settings}                                                     & \multicolumn{2}{c|}{All-search}          & \multicolumn{2}{c|}{Indoor-search} & \multicolumn{2}{c|}{IR-to-VIS}                        & \multicolumn{2}{c}{VIS-to-IR}                        \\ \hline
\multicolumn{1}{c|}{Method}                                           & Publish   & Rank-1        & \multicolumn{1}{c|}{mAP} & Rank-1           & mAP             & \multicolumn{1}{l}{Rank-1} & \multicolumn{1}{l|}{mAP} & \multicolumn{1}{l}{Rank-1} & \multicolumn{1}{l}{mAP} \\ \hline
\multicolumn{1}{c|}{AlignGAN\cite{wang2019rgb}}      & ICCV 19   & 42.4          & 40.7                     & 45.9             & 54.3            & -                          & -                        & -                          & -                       \\
\multicolumn{1}{c|}{DDAG\cite{ye2020dynamic}}        & ECCV 20   & 54.7          & 53.0                     & 61.0             & 67.9            & 40.3                       & 48.4                     & 48.0                       & 52.3                    \\
\multicolumn{1}{c|}{AGW\cite{ye2021deep}}            & TPAMI 21  & 56.5          & 57.4                     & 68.7             & 75.1            & 43.6                       & 51.8                     & 51.5                       & 55.3                    \\
\multicolumn{1}{c|}{MMN\cite{zhang2021towards}}      & ACM MM 21 & 70.6          & 66.9                     & 76.2             & 79.6            & 52.5                       & 58.9                     & 59.9                       & 62.7                    \\
\multicolumn{1}{c|}{CAJ\cite{ye2021channel}}         & CVPR 21   & 69.8          & 66.8                     & 76.2             & 80.3            & 48.8                       & 56.6                     & 56.5                       & 59.8                    \\
\multicolumn{1}{c|}{DART\cite{yang2022learning}}     & CVPR 22   & 60.6          & 58.2                     & 65.7             & 71.7            & 52.2                       & 59.8                     & 60.4                       & 63.2                    \\
\multicolumn{1}{c|}{MSCLNet\cite{zhang2022modality}} & ECCV 22   & 76.9          & 71.6                     & 78.4             & 81.1            & -                          & -                        & -                          & -                       \\
\multicolumn{1}{c|}{PartMix\cite{kim2023partmix}}    & CVPR 23   & 77.7          & 74.6                     & 81.5             & 84.8            & -                          & -                        & -                          & -                       \\
\multicolumn{1}{c|}{SGIEL \cite{feng2023shape}}      & CVPR 23   & 77.1          & 72.3                     & 82.0             & 82.9            & -                          & -                        & -                          & -                       \\
\multicolumn{1}{c|}{DEEN \cite{zhang2023diverse}}    & CVPR 23   & 75.4          & 72.2                     & 82.3             & 84.6            & 54.9                       & 62.9                     & 62.5                       & 65.8                    \\
\multicolumn{1}{c|}{MSCMNet \cite{cheng2023multi}}   & arXiv 23  & 78.5          & 74.2                     & 83.0             & 85.5            & 55.1                       & 60.8                     & 63.9                       & 66.1                    \\
\multicolumn{1}{c|}{HOS-Net \cite{qiu2024high}}      & AAAI 24   & 75.6          & 74.2                     & 84.2             & 86.7            & 56.4                       & 63.2                     & \textbf{64.9}              & \textbf{67.9}           \\
\multicolumn{1}{c|}{PDM}                                              & -         & \textbf{79.3} & \textbf{76.3}            & \textbf{88.7}    & \textbf{89.8}   & \textbf{57.1}              & \textbf{63.6}            & \textbf{64.9}              & 67.3                    \\ \hline
\end{tabular}%
}
\end{table*}

\begin{equation}
    \mathbf{F}_{v} = [\mathbf{p}^i_v, \mathbf{F}^{g}_{v}]
    \label{eq:F_vd}
\end{equation}
where \([\cdot]\) denotes feature concatenation, and \(\mathbf{F}^{g}_{v}\) represents the global feature for the VIS modality. \(\mathbf{F}_{v}\) combines latent semantic similar features and global features. Similarly, this method is applied to \( f^{\textit{ir}} \) to obtain \(\mathbf{F}_{{ir}}\). The learnable prototype facilitates cross-modal semantic alignment. The identity loss \(\mathcal{L}_{id}\) is computed using batch-normalized and classified results derived from \(\mathbf{F}_{v}\) and \(\mathbf{F}_{{ir}}\). Additionally, employing the triplet loss \(\mathcal{L}_{tri}\) supervises the global feature, guiding the model in discerning pedestrian relationships.

\textbf{Cosine Heterogeneity Loss.} 
The Cosine Heterogeneity Loss \(\mathcal{L}_{ch}\) decreases the similarity between each prototypes, thereby enhancing the diversity of information among semantically similar local features extracted by the prototypes. The \(\mathcal{L}_{ch}\) is defined as follows:

\begin{equation}
\mathcal{L}_{ch} = 1 - \frac{2}{m(m-1)} \sum_{i=1}^{m-1} \sum_{j=i+1}^{m} \cos(\mathbf{P}_i \mathbf{I}^\mathsf{T}, \mathbf{P}_j \mathbf{I}^\mathsf{T})
\end{equation}
where $\mathbf{P}_i$ and $\mathbf{P}_j$ denote the $i$-th and $j$-th learnable prototypes, and $\mathbf{I}$ represents $\mathbf{I}_v$ and $\mathbf{I}_{ir}$.

\textbf{Dual-Center Separation Loss.}
We introduce the Dual-Center Separation Loss \(\mathcal{L}_{\textit{dcs}}\) to guide the network in discerning pedestrian relationships. The goal of \(\mathcal{L}_{\textit{dcs}}\) is to draw samples belonging to the same identity closer together while distancing the centers of samples from different identities. We cluster samples within a distance threshold \(\rho_1\) to enhance diversity. The \(\mathcal{L}_{\textit{dcs}}\) is defined as follows:

\begin{equation}
\begin{split}
\mathcal{L}_{dcs} &= \frac{1}{N}\sum^{N}_{i=1} [-\rho_1 + \| \mathbf{F}_i-\mathbf{c}_{y_i} \|_2]_+ \\
&+ \frac{2}{M(M-1)}\sum^{M-1}_{j=1}\sum^{M}_{k=j+1}[\rho_2 - \| \mathbf{c}_{y_j}-\mathbf{c}_{y_k} \|_2]_+
\end{split}
\end{equation}
where \(N\) denotes the batch size, \(\mathbf{F}_i\) represents the \(i\)-th feature, \(y_i\) indicates the \(i\)-th pedestrian, \(\mathbf{c}_{y_i}\) is the centroid of \(y_i\), \(M\) is the number of centroids, \(\rho_1\) signifies the threshold distance from the sample to its centroid and \(\rho_2\) represents the distance between different centroids.

\subsection{Multi-Loss Optimization}

The total loss of the PLM module is as follows:
\begin{equation}
\label{eq:plm loss}
\mathcal{L}_{plm} = \mathcal{L}_{tri} + \mathcal{L}_{ch} + \mathcal{L}_{dcs}
\end{equation}
Besides the \( \mathcal{L}_{cpm} \) and \( \mathcal{L}_{plm} \), we further incorporate \( \mathcal{L}_{id} \)~\cite{ye2021channel} to jointly optimize the network by minimizing these three loss components:
\begin{equation}
\label{eq:total loss}
    \mathcal{L}_{total} = \mathcal{L}_{id} +  \mathcal{L}_{plm} + \mathcal{L}_{cpm}
\end{equation}

\begin{table}[!t]
\centering
 \fontsize{7}{9}\selectfont
\caption{The influence of each component on the performance of the proposed PDM.}
\label{tab:ablation study}
\resizebox{\columnwidth}{!}{%
\begin{tabular}{lcllc|cc}
\hline
 & \multicolumn{4}{c|}{Settings}                                                                                                                         & \multicolumn{2}{c}{SYSU-MM01} \\ \hline
 & PLM                       & $\mathcal{L}_{ch}$                            & $\mathcal{L}_{dcs}$                           & MFGM                      & Rank-1        & mAP           \\
 &                           &                                               &                                               &                           & 64.7          & 62.0          \\
 & \checkmark &                                               &                                               &                           & 71.6          & 66.9          \\
 & \checkmark &                                               &                                               & \checkmark & 73.0          & 70.2          \\
 & \checkmark & \multicolumn{1}{c}{\checkmark} &                                               & \checkmark & 75.7          & 72.2          \\
 & \checkmark &                                               & \multicolumn{1}{c}{\checkmark} & \checkmark & 75.6          & 71.4          \\
 &                           &                                               &                                               & \checkmark & 74.2          & 70.9          \\
 & \checkmark & \multicolumn{1}{c}{\checkmark} & \multicolumn{1}{c}{\checkmark} & \checkmark & \textbf{79.3} & \textbf{76.3} \\ \hline
\end{tabular}%
}
\end{table}

\section{Experiment}

\begin{figure*}[!ht]
    \centering
    \includegraphics[width=2\columnwidth]{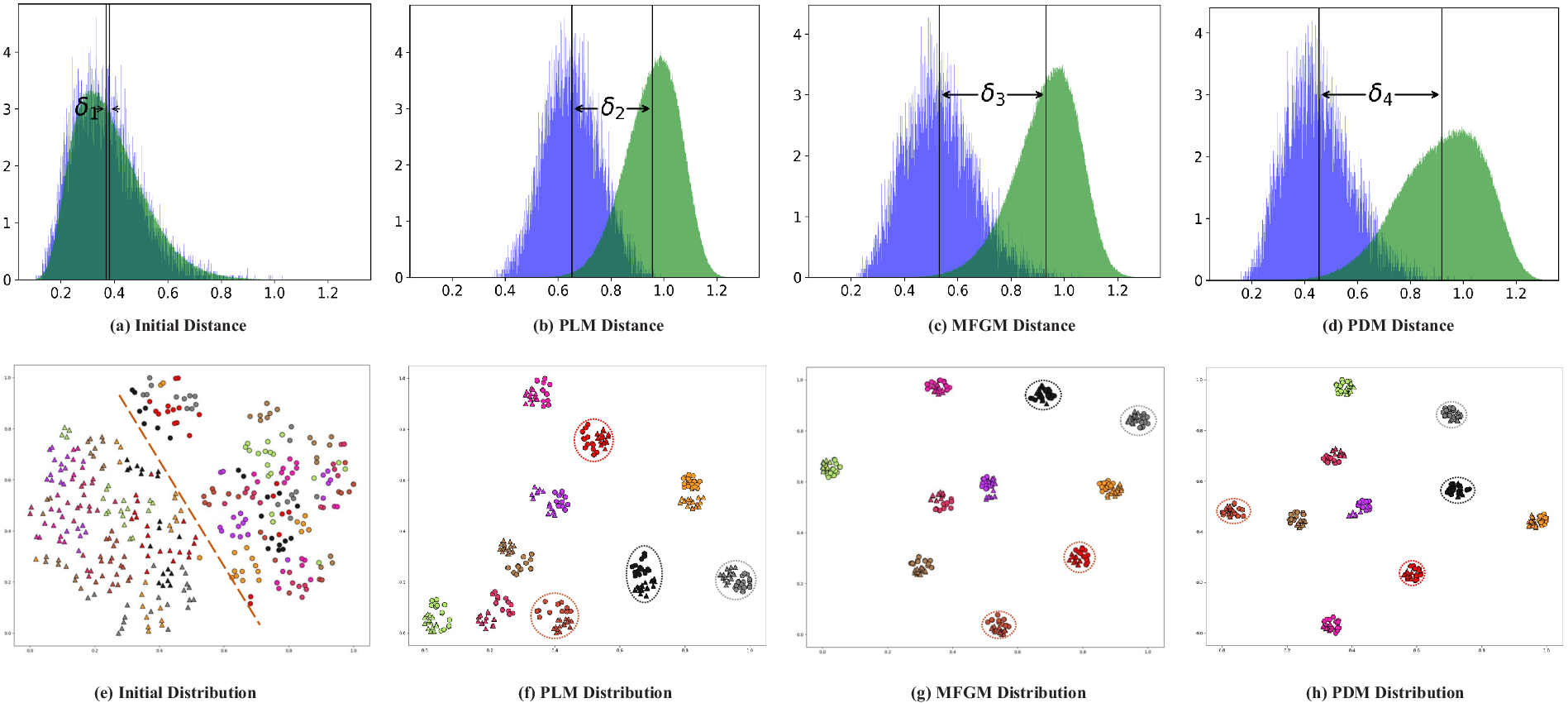}
    \caption{(a-d) illustrate the intra-class and inter-class distances of cross-modality features, with intra-class and inter-class distances represented in blue and green, respectively. In (e-h), the t-SNE~\cite{van2008visualizing} visualizations illustrate the 2D feature distributions, where circles and triangles denote infrared and visible modalities, and different colors represent pedestrians from distinct categories.}
    \label{fig:visulization}
\end{figure*}

\begin{figure}[!t]
    \centering
    \includegraphics[width=\columnwidth]
    {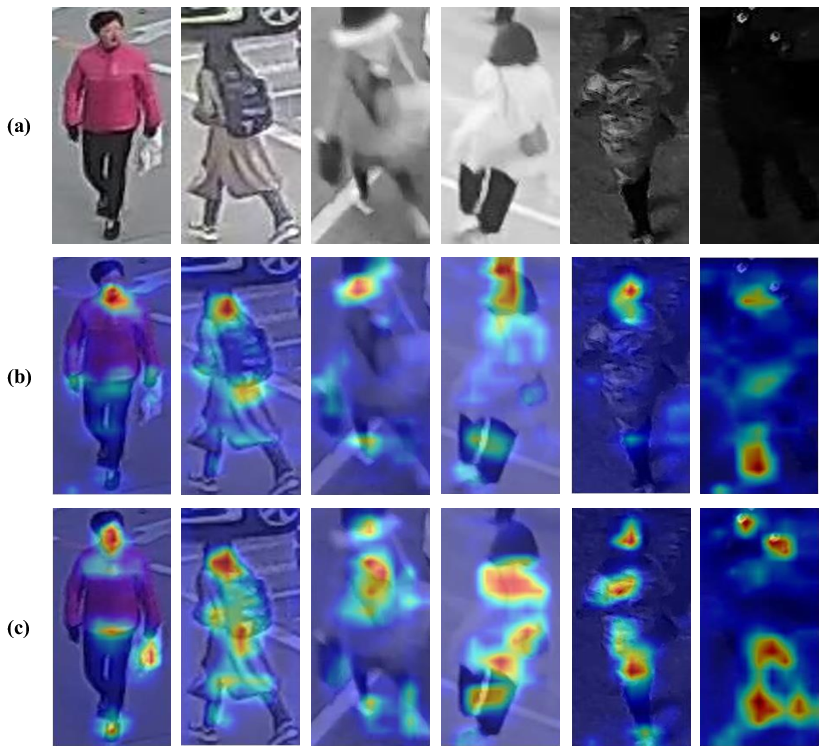}
    \caption{The visualization results of attention maps. (a) represents the displayed image, (b) and (c) show the results of baseline and PDM.}
    \label{fig:attention_map}
\end{figure}

\subsection{Datasets}
We evaluate the performance of our proposed PDM by comparing it with various state-of-the-art methods on the SYSU-MM01\cite{wu2017rgb} and LLCM\cite{zhang2023diverse} datasets.\\
\textbf{Metrics.} In our evaluation, we focus on two pivotal metrics: Cumulative Matching Characteristics (CMC) and Mean Average Precision (mAP).

\subsection{Implementation Details}
The PDM framework is implemented using the PyTorch framework, runs on a single RTX 4090 GPU, utilizing ResNet-50~\cite{he2016deep} as the backbone. Initial input images are resized to a consistent dimension of \(3 \times 384 \times 192\). Various augmentation techniques are applied, including random horizontal flipping and random erasing. The initial learning rate is set to \(1 \times 10^{-2}\) and increased to \(1 \times 10^{-1}\) after 10 epochs. Subsequently, at 80 and 120 epochs, it undergoes further decay to \(1 \times 10^{-3}\) and \(1 \times 10^{-4}\), respectively, concluding a total training period of 150 epochs. The training process employs the SGD optimizer with a momentum of 0.9. Additionally, we set the number of learnable prototypes \(m\) to 10.

\subsection{Main Results}
As shown in Table~\ref{tab:dbsy}, PDM outperforms competing methods in cross-modality person re-identification tasks. On the SYSU-MM01 dataset, it achieves a rank-1 accuracy of 79.3\% and mAP of 76.2\% in the All-search mode, and 88.7\% rank-1 accuracy and 89.8\% mAP in the Indoor-search mode. On the LLCM dataset, PDM achieves a rank-1 accuracy of 57.1\% and mAP of 63.6\% in the IR-to-VIS mode, and 64.9\% rank-1 accuracy and 67.3\% mAP in the VIS-to-IR mode. These results demonstrate PDM's effectiveness in addressing modality disparities and its exceptional performance in cross-modality person re-identification tasks. Additionally, on the SYSU-MM01 dataset, PDM surpasses HOS-Net with a 3.7\% higher rank-1 accuracy and 2.1\% higher mAP. In the LLCM dataset, PDM outperforms HOS-Net by 0.7\% in the IR-to-VIS mode and exhibits a slightly lower mAP by 0.6\% in the VIS-to-IR mode. This underscores PDM's superior performance and effectiveness in handling modality disparities.

\subsection{Ablation Studies}
\textbf{Effectiveness of each component.} 
The ablation studies conducted on the SYSU-MM01 dataset, as presented in Table~\ref{tab:ablation study}, demonstrate the effectiveness of PLM and MFGM components individually and in combination. Including \(\mathcal{L}_{ch}\) and \(\mathcal{L}_{dcs}\) enhances the model to achieve optimal performance.

\textbf{Effectiveness of different numbers of learnable prototypes for the PLM.}
The PLM utilizes learnable prototypes to discover semantically similar local features across modalities. Our study explores different numbers of prototypes for the PLM and finds that performance improves as the number increases from 6 to 10. However, as shown in Table~\ref{tab:proto abl}, performance starts to decline beyond 10 prototypes. Setting the number to 10 achieves the best performance on the SYSU-MM01 dataset, leading us to adopt this configuration for the PLM.

\begin{table}[!t]
\centering
 \fontsize{4}{7}\selectfont
\caption{The influence of different quantities of learnable prototypes on the performance of the proposed PDM.}
\label{tab:proto abl}
\resizebox{\columnwidth}{!}{%
\begin{tabular}{c|cccc}
\hline
\multirow{2}{*}{\centering Settings} & \multicolumn{2}{c|}{\centering All-search} & \multicolumn{2}{c}{\centering Indoor-search} \\ \cline{2-5} 
         & Rank-1        & \multicolumn{1}{c|}{mAP} & Rank-1        & mAP           \\ \hline
m = 6  & 78.4          & 75.2                     & 86.5          & 88.3          \\
m = 8  & 78.6          & 75.6                     & 85.8          & 87.8          \\
m = 10 & \textbf{79.3} & \textbf{76.3}            & \textbf{88.7} & \textbf{89.8} \\
m = 12 & 78.1          & 75.8                     & 85.2          & 87.1          \\ \hline
\end{tabular}%
}
\end{table}

\subsection{Visualization Analysis}
\textbf{Feature Distribution.}
We conducted an analysis of intra-class and inter-class distance distributions for cross-modality features on the SYSU-MM01 dataset, as depicted in Fig.~\ref{fig:visulization} (a-d). The mean values, indicated by vertical lines, exhibit a progressive divergence (\(\delta_1 < \delta_2 < \delta_3 < \delta_4\)). By integrating PLM, we observed an increase in the inter-class distance and an enlargement of the gap between the average intra-class distance and inter-class distance. Furthermore, with the incorporation of MFGM, the intra-class distance decreased, leading to a further enhancement of the gap. Notably, the combination of both modules resulted in the maximum gap. To visually demonstrate the discriminative capability of the PLM, MFGM, and PDM, we conducted t-SNE visualizations (Fig.~\ref{fig:visulization} (e-h)), which illustrated the clustering of embeddings per individual. These visualizations reaffirm that the PDM (Prototype Distribution Mining) approach effectively addresses intra-modal and inter-modal disparities in cross-modal person re-identification. By leveraging diverse features that exhibit close distributions and utilizing learnable prototypes to capture latent semantic similarities among cross-modal features, PDM enables a joint representation of pedestrians using multiple partial features, effectively mitigating both intra-modal and inter-modal variations. These comprehensive analyses consistently validate the efficiency of our proposed method in the context of cross-modality person re-identification.

\textbf{Attention Visualization.} 
Figure~\ref{fig:attention_map} illustrates attention maps, showing that PDM focuses more on pedestrian regions compared to the baseline method.  These analyses validate the effectiveness of PDM in mitigating inter-modal disparities and capturing semantic similarities among cross-modal features.

\section{Conclusion}
We propose PDM, a Prototype-Driven Multi-Feature Generation Network for cross-modal person re-identification. PDM consists of two modules: Multi-Feature Generation Module (MFGM) and Prototype Learning Module (PLM). MFGM extracts diverse features from modality-specific inputs to enhance shared information, aligning their distributions with a center-guided pair mining loss. PLM integrates learnable prototypes to weight modality-specific features based on prototype similarity, facilitating the discovery of semantically similar local features across modalities for fine-grained alignment. By combining local and diverse features, PDM effectively mitigates inter-modal and intra-modal discrepancies. Experimental results on SYSU-MM01 and LLCM datasets demonstrate PDM's state-of-the-art performance in person re-identification. 

In the future work, we will focus several directions to improve VI-ReID: (1) applying more advanced attention-based feature aggregation mechanism~\cite{yang2020gated} for better representation learning ; (2) adopting contrastive learning~\cite{yang2022mutual,yang2023online} to enhance the discriminative ability; (3) introducing CLIP~\cite{radford2021learning,yang2024clip} to promote multi-modality information processing; (4) combining knowledge distillation~\cite{yang2021hierarchical,yang2022cross,yang2022mixskd,feng2024relational} for VI-ReID model compression. 

\bibliographystyle{IEEEtran}
\bibliography{IEEEabrv, refs}


\end{document}